\documentclass[conference]{IEEEtran}
\IEEEoverridecommandlockouts

\usepackage{cite}
\usepackage{amsmath,amssymb,amsfonts}
\usepackage{algorithmic}
\usepackage{graphicx}
\usepackage{textcomp}
\usepackage{xcolor}
\usepackage{subfig}
\usepackage{svg}
\usepackage{colortbl}
\usepackage{array}
\usepackage{comment}
\usepackage{soul}

\def\BibTeX{{\rm B\kern-.05em{\sc i\kern-.025em b}\kern-.08em
    T\kern-.1667em\lower.7ex\hbox{E}\kern-.125emX}}
\begin{document}

\title{Task-oriented grasping for dexterous robots using
postural synergies and reinforcement learning
\thanks{
Work  partially supported by 
the H2020 FET-Open project \textit{Reconstructing the Past: Artificial
Intelligence and Robotics Meet Cultural Heritage} (RePAIR) under EU grant
agreement 964854, by the Lisbon Ellis Unit (LUMLIS),
by the FCT PhD grant [PD/BD/09714/2020], by LARSyS FCT funding (DOI: 10.54499/LA/P/0083/2020, 10.54499/UIDP/50009/2020, and 10.54499/UIDB/50009/2020) and and by the Portuguese Recovery and Resilience Plan (RRP), project number 62, Center for Responsible AI. }
}

\author{\IEEEauthorblockN{Dimitris Dimou}
\IEEEauthorblockA{\textit{Institute for Systems and Robotics} \\
\textit{Instituto Superior Tecnico}\\
Lisboa, Portugal\\
mijuomij@gmail.com}
\and
\IEEEauthorblockN{Jos\'e Santos Victor}
\IEEEauthorblockA{\textit{Institute for Systems and Robotics} \\
\textit{Instituto Superior Tecnico}\\
Lisboa, Portugal\\
jasv@isr.tecnico.ulisboa.pt}
\and
\IEEEauthorblockN{Plinio Moreno}
\IEEEauthorblockA{\textit{Institute for Systems and Robotics} \\
\textit{Instituto Superior Tecnico}\\
Lisboa, Portugal\\
plinio@isr.tecnico.ulisboa.pt}
}

\maketitle

\begin{abstract}

In this paper, we address the problem of task-oriented grasping
for humanoid robots, emphasizing the need to align with human
social norms and task-specific objectives. Existing methods, 
employ a variety of open-loop and closed-loop approaches but 
lack an end-to-end solution that can grasp several objects
while taking into account the downstream task's constraints.
Our proposed approach employs reinforcement learning to enhance 
task-oriented grasping, prioritizing the post-grasp intention 
of the agent. We extract human grasp preferences from the 
ContactPose \cite{Brahmbhatt_2020_ECCV} dataset, and train a 
hand synergy model based on the Variational Autoencoder (VAE) 
to imitate the participant's grasping actions. Based on this 
data, we train an agent able to grasp multiple objects while 
taking into account distinct post-grasp intentions that 
are task-specific. By combining data-driven insights from
human grasping behavior with learning by exploration provided 
by reinforcement learning, we can develop humanoid robots 
capable of context-aware manipulation actions, facilitating 
collaboration in human-centered environments.
\end{abstract}


\section{Introduction}


Human grasping behavior is influenced by the broader context of 
the manipulation action to be performed. One important factor 
that humans take into account for grasping an object is 
the reason behind grasping it in the first place. A common 
 example is when we want to grasp something in
order to use it or to hand it over to another person. 
In this case we tend to grasp the object from its functional
part when we intend to use it, while we leave that part free
when we want to hand it over, this way the other person can 
grasp it from its functional part and use it directly without
needing to perform a regrasp. 

Having humanoid robots that follow human social norms is 
important to accelerate human-robot collaboration and
facilitate the introduction of humanoids to human centered
environments. To achieve this we need to develop grasping 
methods that can imitate the human grasping behavior. This
means that the robot must take into account task-specific 
factors, enabling more sophisticated and context-aware 
manipulation actions. The primary challenge of this problem 
lies in its inherent connection with human behavior, 
preferences, and social norms. Therefore, any proposed 
solutions must be derived from real-world data that 
encapsulate these characteristics.

\begin{figure}[h] \centering
\includegraphics[width=0.23\textwidth]{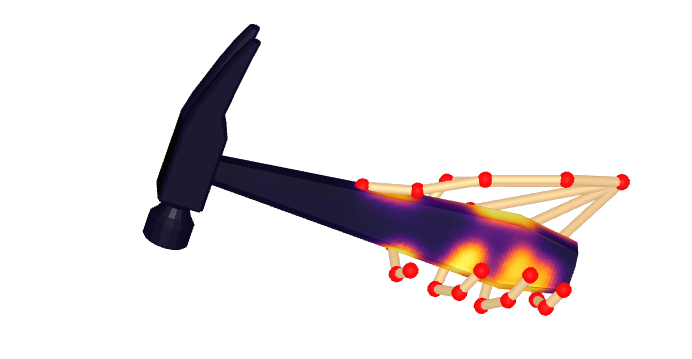}
\includegraphics[width=0.23\textwidth]{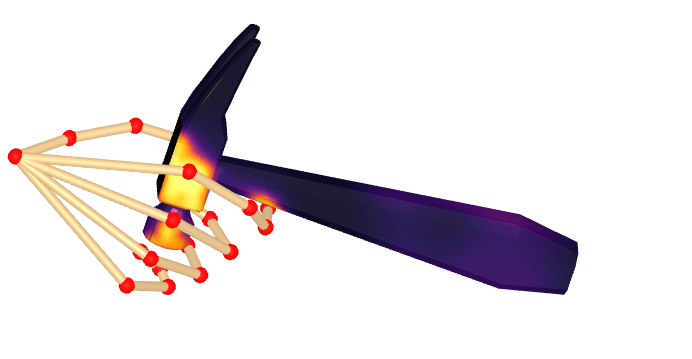}
\caption{\label{fig:task_example} 
Example of different execution of a grasp according to the post-grasp
intention as captured in the dataset presented in \cite{Brahmbhatt_2020_ECCV}.
In the left figure, the person grasps the hammer in order to use it, in
the right it grasps it in order to hand it over to another person. }
\end{figure}

Previous works have employed open-loop grasp generation methods,
where a grasp sampling method produces a static grasp pose, 
considering task constraints to ensure alignment with the task’s 
function. Subsequently, a motion planning algorithm devises a 
trajectory to execute the proposed grasp. Conversely, in closed-loop
approaches, a policy utilizes observations and task constraints 
as inputs, generating actions that result in suitable object 
grasping. Open-loop methods often depend on extensive labeled 
datasets for training the grasp generation modules. However, 
their inherent nature prevents them from adjusting trajectories 
based on real-time observations or accommodating measurement 
uncertainties. In contrast, closed-loop methods typically 
leverage reinforcement learning algorithms to train grasping 
policies through exploration, necessitating numerous training 
samples and complex reward shaping to incorporate human priors.

In this work, we explore task-oriented grasping using 
reinforcement learning based on the intended post-grasp 
manipulation action. We develop an agent that can grasp 
objects in a way that is well-suited for the particular 
task that wants to accomplish. To achieve this, we 
start by extracting the grasp preferences from the ContactPose
\cite{Brahmbhatt_2020_ECCV} data, which contains human grasps
on various objects with the intention to use them or to
hand them over.
We retarget the human grasps postures to the robotic hand 
and use them to train a synergy model, based on the VAE model,
to generate new hand postures in a low-dimensional space. 
We, then, train an agent (i.e. a single policy), using 
reinforcement learning, that uses the synergy model as 
action space, to grasp multiple objects conditioned on 
a specified post-grasp intention.

To sum up our main contributions are the following: 
\begin{itemize}
    \item We train one policy to grasp multiple objects
    while taking into account the post-grasp intention of the agent.

    \item We utilize a hand synergy model trained on grasps 
    performed by humans to improve the quantitative and 
    qualitative performance of the agent. 

    \item We demonstrate that the agent is able to grasp objects
    based on it's post-grasp intention similarly to how humans
    grasped them.
\end{itemize}

\section{RELATED WORK}
Previous works on task-oriented grasping can be categorized into
open-loop grasp pose generation methods and closed-loop grasping
approaches. Typically, open-loop approaches divide the grasping 
behavior into two stages: first, a grasp pose is generated, usually
represented by a 6DoF pose for the hand and finger configuration 
and then the target grasp pose is passed to a motion planning 
algorithm to find a feasible trajectory for execution. In this 
case, the our interest lies in the grasp pose generation phase. 
In \cite{Song2011LearningTC}, they use a classic geometric grasp
sampling method from the GraspIt! simulator to generate grasps 
for several household objects. They then label the grasp-object 
pairs with task descriptions, such as hand-over, pouring, or 
tool-use. Finally, they train a Bayesian Neural Network, to model
the relationships between the recorded features. During runtime, 
they query the network to output grasp poses given the task label
and object properties. In \cite{Li2016LearningPP}, they train a
Gaussian Mixture Model to model the correlations between task-specific
grasps and object properties. New grasps are then predicted using 
Gaussian Mixture Regression based on the learned model.
In a more recent study \cite{Murali2020SameOD}, they develop a Graph
Convolutional Neural Network (GCN) to encode grasp, object, and task
relationships. When presented with a new object, multiple grasp poses
were sampled using a geometric heuristic, and the grasp-object pair 
was encoded into the GCN alongside the task description. The grasp 
evaluator then predicted a grasp score, and the grasp with the highest
score was selected.

Another line of work involves visually detecting affordances on 
the objects, i.e. parts of the object that are compatible with 
a specific task and then using a  grasp sampling method to estimate
suitable grasps for that part.
For example, in \cite{Kokic2017AffordanceDF}, they use a 
Convolutional Neural Network to detect task affordance regions
on objects from point clouds. Given a task and a voxelized
point cloud of an object, the network predicts which voxels 
of the object are compatible with the given task. Given
the graspable part of the object, an optimization-based grasp
sampler computes contact points that result in a high-quality
grasp. In \cite{Liu2020CAGECG}, they develop a discriminative 
model that predict a label indicating whether a grasp is suitable 
for a given context. They use a neural network architecture to
detect affordances for each part of the object. A geometric 
heuristic then samples new grasps predict a score. 
The grasp with the highest score was finally selected.

Open-loop grasp methods offer a modular approach to grasp 
sampling and can generate of multiple candidate grasps for 
a given context. However, a significant drawback is that 
training such models typically requires large amounts of 
labeled data, particularly with human-based annotations, 
which can be challenging to obtain. Additionally, incorporating
trajectory planning and collision avoidance adds another 
layer of complexity. Notably, open-loop grasp generation 
with task constraints has not yet been widely applied to 
humanoid hands, possibly due to their increased complexity 
compared to simpler grippers or robotic hands.

Closed-loop methods offer an alternative to open-loop grasp
generation by modeling the entire grasping action. Typically,
these methods involve training a policy using reinforcement
learning. The key challenge lies in appropriately defining the
task and designing a suitable reward function to guide the 
policy towards desired behavior.

For example, in \cite{Mandikal2021LearningDG}, they used data 
from \cite{ContactDB}, which consists of 3D models of everyday
objects annotated with contact maps captured from humans 
performing two tasks (using the object and handing it off). 
Using this data, they trained a CNN to detect affordance 
regions on objects from RGB images. The affordance map, along
with the state of a humanoid robotic hand, formed the state
representation in a reinforcement learning environment. 
The reward function encouraged the agent to grasp the objects
from the same affordance regions that human subjects did.
However, the agent's resulting finger configurations were
unnatural and did not resemble those of a human hand.
To address this limitation and encourage more human-like grasps,
the authors later extended their work in \cite{Mandikal2021DexVIPLD}.
They introduced an auxiliary reward that measures the distance
between the robotic hand's posture and a human hand posture 
grasping the same object. Human hand poses were extracted from
internet videos using a hand pose estimation algorithm. This 
addition led the new agent to learn more natural hand movements, 
resulting in an increased grasp success rate. However both approaches
focus on grasping the objects from their functional part, i.e. 
in order to use it, without taking into account other post-grasp
intentions of the agent.


Reinforcement learning approaches offer certain advantages, as 
they do not require successful grasp examples for training, 
relying instead on learning from experience. Additionally, these
policies can perform the entire grasping task, eliminating 
the need for separate trajectory planning or collision avoidance
steps. Their closed-loop nature also makes them more robust in
the face of external perturbations or measurement uncertainties.
However, a crucial limitation lies in their dependency on specific
reward functions, potentially restricting their applicability to 
only the tasks defined by these rewards. Consequently, using the 
grasping behavior in a new task may require retraining the agent 
with a different reward function.
On the other hand, these algorithms demand vast amounts of samples, 
making simulation-based training more feasible than real-world
applications. Additionally, effective training often requires 
careful reward shaping to achieve desirable behavior.



\section{METHODS}
The goal of this work is to develop an agent that can grasp
several objects based on the post-grasp intention that will be
given as a conditional variable to the policy. We assume that
the policy has access to the global pose of the object and the 
robot's state. The goal of the agent is to lift the corresponding 
object from the table in front of it while taking into account 
its post-grasp intention. To achieve this we extract grasp data
from a dataset of human grasp with different post-grasp intentions
and then train a policy with reinforcement learning  to exhibit
the desired behaviour.

\begin{figure}[t] \centering
\includegraphics[width=0.2\textwidth]{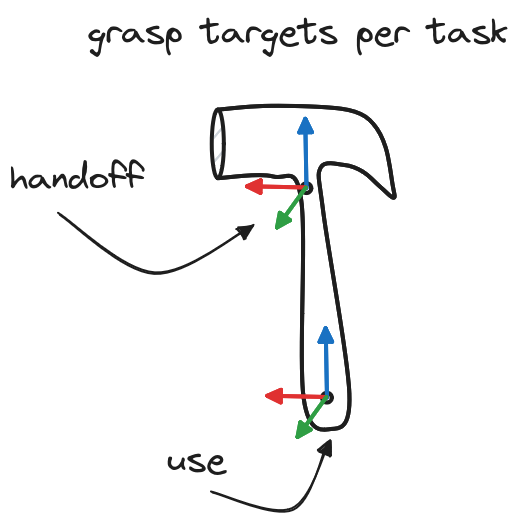}
\caption{\label{fig:hammer_grasp_targets} 
Example grasping targets for hammer extracted from the 
ContactPose dataset \cite{Brahmbhatt_2020_ECCV}.}
\end{figure}

\subsection{Dataset pre-processing} 

We rely on the dataset introduced in \cite{Brahmbhatt_2020_ECCV}
which consists of 3D models of everyday objects annotated with
contact maps captured from humans operators grapsing the objects
with two different post grasp intentions: 
1) using the object and 2) handing the object off
to another person. The dataset also contains the hand posture,
i.e. the 3D position of keypoints on the fingers of each subject,
which was estimated through a 3D vision pipeline.
We use two types of data in our work: 1) the grasp postures of
the human subjects and 2) the grasp locations that they chose.

\textbf{Hand postures.}
Firstly, we retarget the grasp postures executed on the objects
by the participants to the robotic hand by designing a fixed
mapping function between the two kinematic chains. 
More specifically, for each joint we compute the angle defined by
the adjacent points (red points in Figure \ref{fig:task_example})
and we remap this angle to the joint limits of the robotic hand.
We perform this procedure for all grasps and all objects in the
ContactPose dataset, resulting in $2596$ grasps.
Using this dataset we train a VAE model to extract the synergy
space following the method presented in \cite{9560818}.
This model allows us
to sample new grasps by providing new latent points.

\textbf{Grasp targets.} \label{grasp_targets}
Subsequently, we determine target grasp points on each object 
that correspond to each post-grasp intention. Example 
grasp target points on the hammer, according to Figure 
\ref{fig:task_example}, can be seen in Figure \ref{fig:hammer_grasp_targets}. 
To automate this process, for each grasp we compute a 3D point
that is the average position of the fingertips of the thumb, the 
index, and the middle finger. Each object was grasped by multiple
participants, so for each object we have multiple 3D points 
indicating the location where each participant grasped it from.
Since the dataset includes two post-grasp intentions, we cluster 
the contact points into two clusters. For each cluster then we
check if there were more handoff grasps or use grasps and assign
the corresponding label to it. The center of the cluster is then
defined as the grasp target for that object and that post-grasp
intention.
These target grasp points are used to guide the reinforcement
learning policy to grasp the object close to it.
Given the synergy grasp model and the grasp targets for each
object and post-grasp intention we can move on to train our
policy.

\begin{figure}[t] \centering
\includegraphics[width=0.45\textwidth]{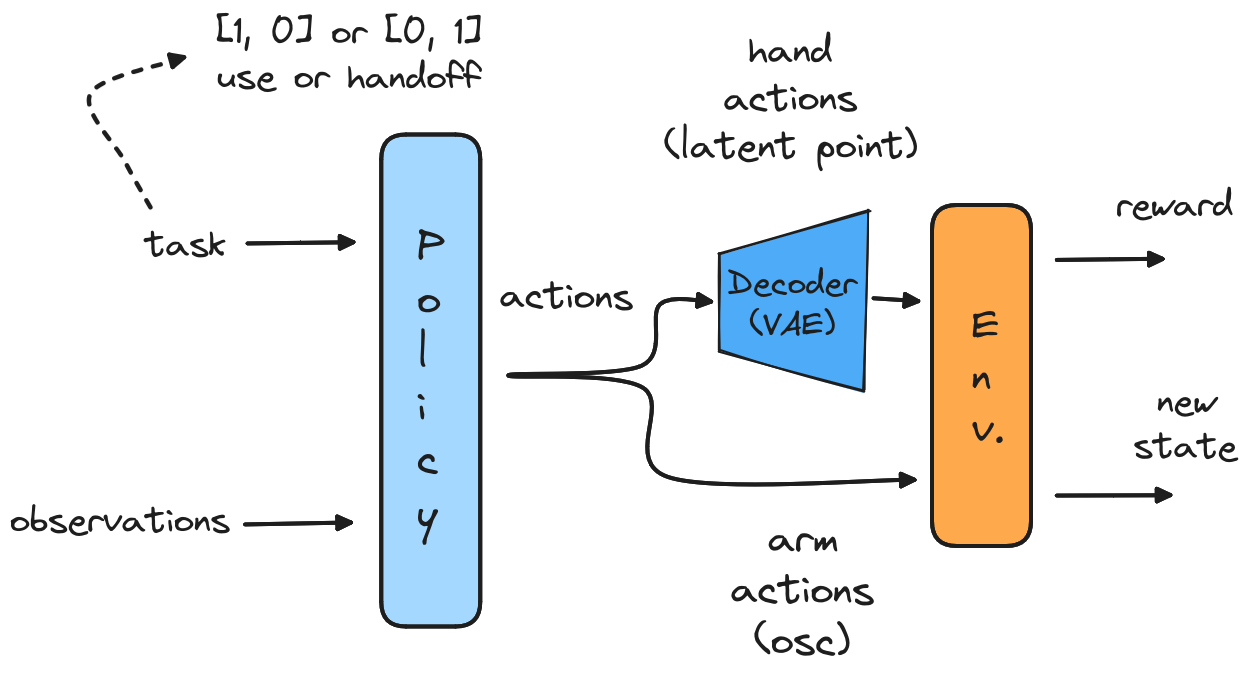}
\caption{\label{fig:tog_policy} 
Proposed agent structure for task-oriented grasping.}
\end{figure}

\subsection{Dexterous task-oriented grasping}
Our goal is to develop an agent that can grasp multiple objects
based on the post-grasp intention that it is conditioned on.
To this end, we train a policy using reinforcement learning
to generate the required actions to perform the task.
We assume that the policy has access to the pose of the objects,
the target object, and proprioceptive information, which are
joined together in the observation variable. We also assume 
that the post-grasp intention is given as a conditional variable
to the policy. 

\textbf{Policy.}
A graphical representation of the policy structure can be 
seen in Figure \ref{fig:tog_policy}. The policy is modeled 
with a neural network that takes as input the current 
observations and the post-grasp intention and outputs two 
actions, one for the hand and one for the arm. 
The hand action is a low-dimensional point in the
synergy space which is then decoded by the synergy model into
the finger joint values. The arm action is the end-effector
displacement in Cartesian space which is used by an Operational
Space Controller (OSC) \cite{OSC} to compute the desired joint angles.
To optimize the policy we use the Proximal Policy Optimization 
(PPO) \cite{schulman2017proximal} algorithm which is a popular 
online reinforcement learning algorithm. Note that we train
one policy that can handle all objects and post-grasp intentions.

\textbf{Action and state space.}
The policy takes as observations the robot's state, i.e. 
the joint angle values, the object's state,
i.e. its 6DoF pose, and its category, which is an one-hot
encoded variable. In addition, it is given a task-conditional variable, 
that indicates the post-grasp intent, which is also one-hot encoded. 
Consequently, the policy generates actions for the robot to execute.
The actions are a 6 DoF pose for the end-effector that is commanded 
to the robotic arm using an inverse kinematics controller, and a latent
point that is used by the VAE model to generate the grasp posture of 
the robotic hand. 

\textbf{Reward function.}
The policy is trained to lift the specified object from the table
by taking into account the post-grasp intent. To achieve this, we
design the reward function such that the reward is high when the 
object is lifted and the grasp location is close to the desired 
grasp target. Our reward function depends only on the current state
and is formulated as follows:
$$
r = w_1 * r_{hand\_obj\_dist} + w_2 * r_{lift} + w_3 * r_{rotation} 
$$
$$
r_{hand\_obj\_dist} = r_{fingertips\_object\_dist} + r_{palm\_object\_dist}
$$
$$
r_{lift} = r_{object\_height} + r_{object\_grasped}
$$
The reward is the sum of three terms that 
are used to achieve specific outcomes. Firstly, the reward measures
the proximity of the grasp location associated with the desired 
grasp target. The $r_{hand\_obj\_dist}$ function measures the 
distance from the fingertips to the target grasp location, 
${fingertips\_object_dist}$, and the distance from the palm to the 
target grasp location on the object, $r_{palm\_object\_dist}$.
As the distance get smaller the reward increases. This encourages
the policy to place the hand on the target grasp location and close the 
fingers.
By reinforcing a high reward for a close alignment 
between the chosen grasp location and the specified target, the 
learning process is fine-tuned to prioritize precision in object 
manipulation.
Secondly, the reward is structured to attain a high
value upon successful lifting of the targeted object. This encourages 
the agent to converge towards actions that lead to the successful 
execution of the lifting task. 
The $r_{lift}$ function consists of two terms 
one that measures the height of the object, $r_{object\_height}$
and increases as the object height increases, and a binary term,
$r_{object\_grasped}$ that gives a positive reward if the object
is lifted above a specific height.
Finally, we include a rotation reward, $r_{rotation}$, for the hand
that encourages the hand to keep its orientation towards the floor, 
and facilitate the exploration phase.


\section{EXPERIMENTS}

\begin{figure*}[t] \centering
\includegraphics[width=0.8\textwidth]{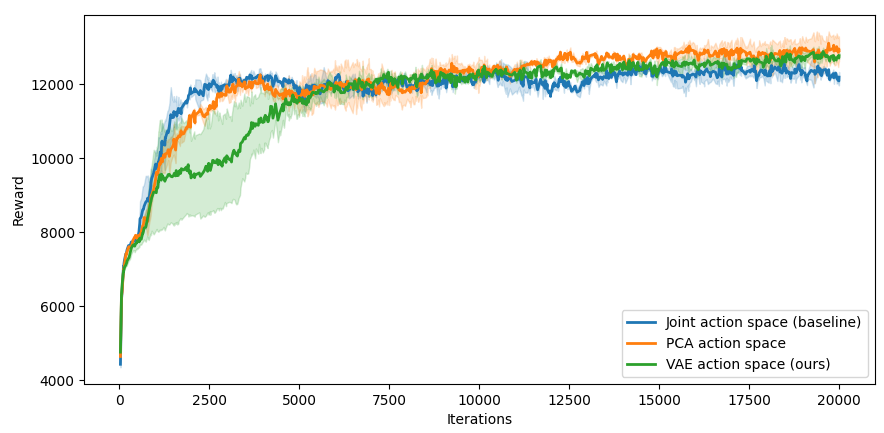}
\caption{\label{fig:pol_rew} 
Rewards for training policies with 1) full joint control, 
2) PCA synergy space, and 3) VAE synergy space.
The thick line is the average 
among the two seeds and the shaded part denotes the standard 
deviation.}
\end{figure*}

\subsection{Experimental set-up}
\begin{figure}[h] \centering
\includegraphics[width=0.4\textwidth]{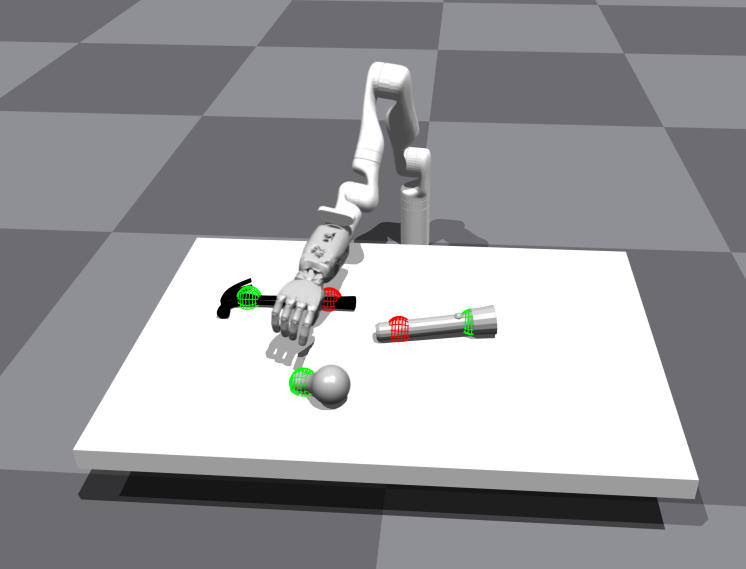}
\caption{\label{fig:env} 
Training environment.}
\end{figure}

\textbf{Implementation details.}
For all our experiments we use the 7 DoF Kinova Gen3 robotic arm 
with the Seed Robotics hand, which in simulation has 19 DoFs. 
For simulation we use Nvidia's IsaacGym physics based simulator
that allows to train multiple environments in parallel. 
From the ContactPose database we choose the following objects:
hammer, flashlight, and light bulb. We chose these objects 
due to the fact that the post-grasp intentions were clearly 
separated on the object's surface. For example, the hammer
all grasps associated with the \textit{use} post-grasp intention
were on the handle and most of the grasps associated with
the \textit{use} post-grasp intention were on the head. In contrast,
the grasps on the other objects were more randomly placed on the
objects making it hard to separate the use cases.

The policy network consists of an RNN network with 768 
hidden units and 1 layer 
followed by an MLP network with 3 layers with 768, 512, 256 units
respectively and ELU activation functions. Each policy was trained
with 2048 parallel environments, for 20k epochs and for
2 different seeds. In our simulated environment the robot is
placed on the floor and all the available objects on a table 
in front of it (Figure \ref{fig:env}).
The object's position and rotation are randomized
using additive Gaussian noise.

\textbf{Evaluation.}
We evaluated our approach by comparing it to a 
reinforcement learning (RL) policy that directly 
outputs finger joint values to control the hand. 
The baseline policy was designed with identical 
observations, reward functions, and action space 
for the arm, and utilized the same network architecture
as our model. Additionally, we compared against a 
policy that utilizes a synergy model based on the 
principal component analysis (PCA) method.
PCA is a popular approach to model the
synergy space of robotic hands and has been used in several
works \cite{EigenGrasp,Bernardino}.
For the evaluation, we conducted 5000
environment simulations to calculate the average 
grasp success rate. Additionally, we documented 
the rewards achieved during training and the final
distances from the hand to the grasp targets for
each task. We further examined the impact of 
increasing the number of latent dimensions in 
the synergy model on policy performance. Lastly, 
we performed an ablation study by excluding the 
object category from the policy's observations 
to assess its influence.

\subsection{Results}
\begin{figure}[h] \centering
\includegraphics[width=0.45\textwidth]{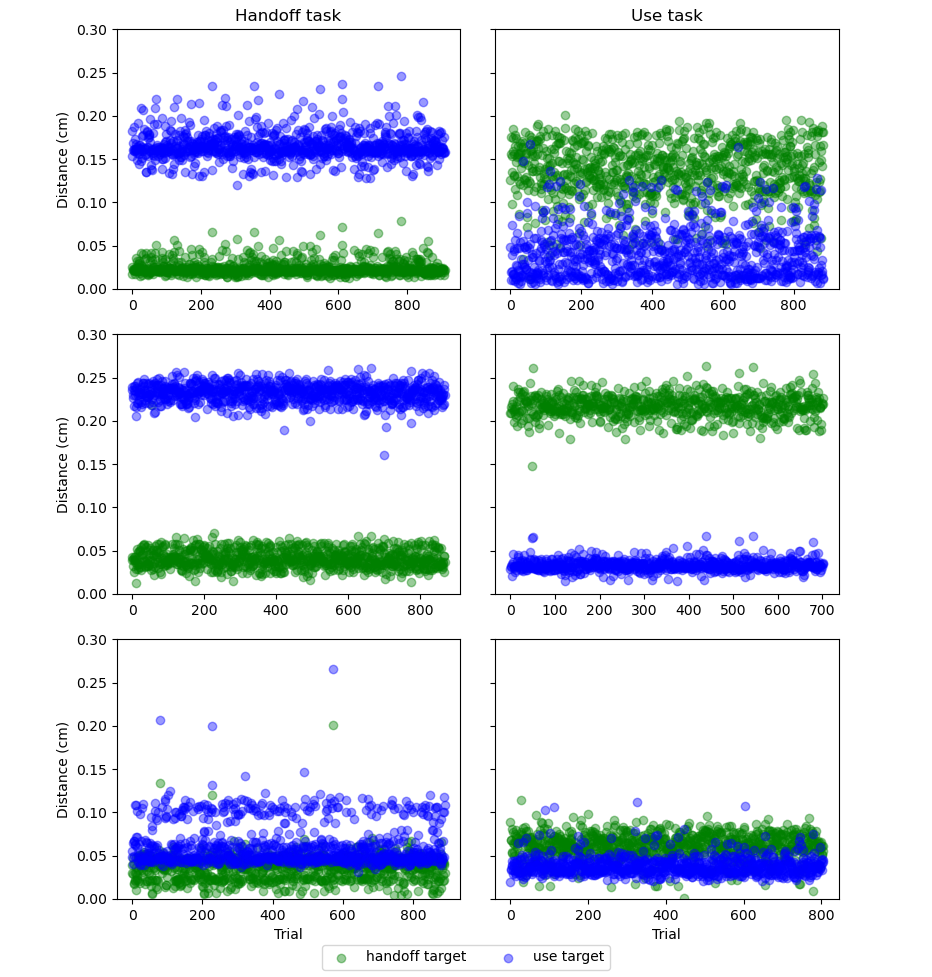}
\caption{\label{fig:grasp_dists} 
Distances of the robot's hand to each grasp target for each task. 
The results are for a 1000 grasp trials performed for each object 
and each post-grasp intention performed using our proposed approach.}
\end{figure}

\textbf{Quantitative.}
Figure \ref{fig:pol_rew} shows the accumulated rewards that each
policy achieved during training. The policy that uses the joint action space 
achieves higher average rewards in general which is probably 
because it learns to grasp the object faster and has lower 
variance across the different seeds. On the other hand, the policy
with the PCA action space achieves the highest final rewards. 
The accumulated reward though is not binary, i.e.
if the object was grasped or not, but depends on how
fast the object is grasped and how high the object is lifted, so
it is not a good proxy for the grasping performance of the agent.
Since the reward alone does not guarantee better performance 
in the task at hand, we compared the best policy from each method 
using the grasp success rate as a metric. More specifically, we run
5000 test environments using the best policies and computed the average 
grasp success rate, measured by the times the object was lifted 
from the table in each episode. The results, shown in Table 
\ref{table:success_rate}, indicate that the policy with the VAE
action space perform better in terms of grasp success, which is
the objective of the agent. 
Moreover, in Figure \ref{fig:grasp_dists}, we confirm the correct
behavior of the policy according to the task. To determine that, 
we show the distance of the robot's hand to each grasp target 
(which was defined for each object as described in \ref{grasp_targets})
for the two post-grasp intentions. This distance was computed at 
the end of each episode, so at the final position of the hand with 
respect to the object. We performed $1000$ trials for each object 
and each post-grasp intention and kept the results of the trials 
where the grasp was successful. For example, the right column of 
figures, which correspond to trials with the handoff post-grasp 
intention, shows that in every successful trial the distance of 
the hand to the handoff target (green dots) defined on the object 
was lower than the distance of the hand to use grasp target. On 
the other hand, on the left figures, which correspond to the use 
post-grasp intention, the distances are the opposite. This demonstrates
that the policy actually learns to associate each grasp position on
the object to a different post-grasp 
intention. 

\begin{figure}[h] \centering
\includegraphics[width=0.49\textwidth]{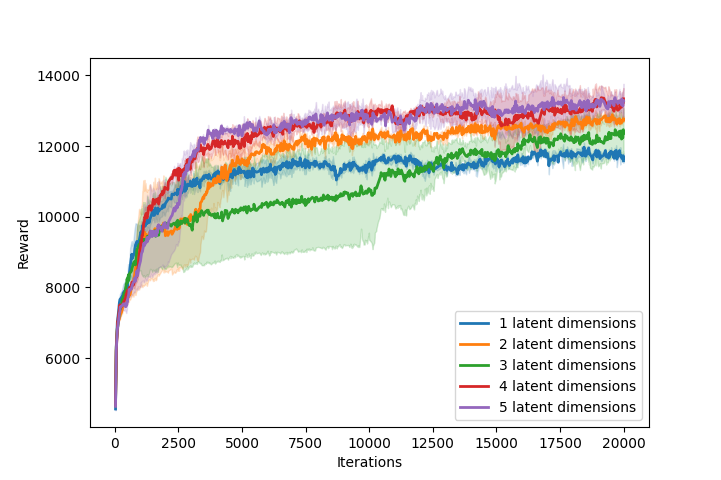}
\caption{ 
Rewards for policies that use synergy models of different latent dimensions.}
\label{fig:rewards_per_latdim}
\end{figure}

\begin{figure}[h] \centering
\includegraphics[width=0.45\textwidth]{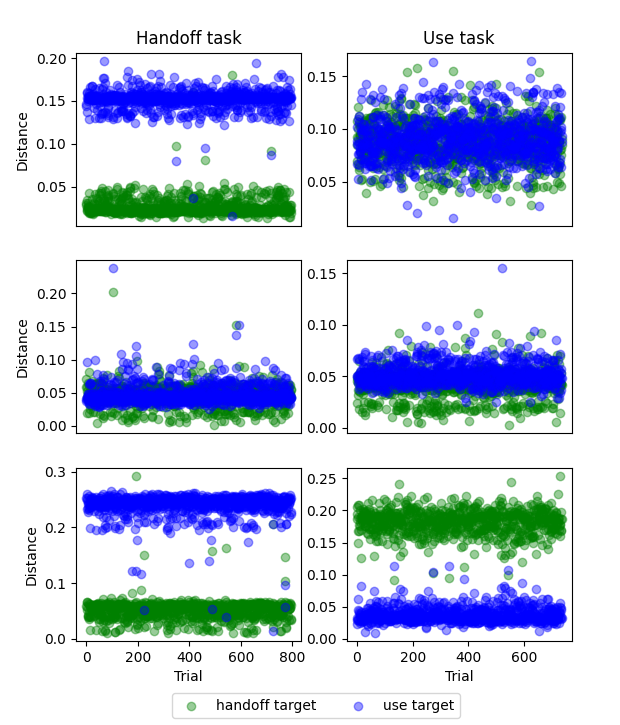}
\caption{\label{fig:distances_no_object_cat} 
Distances of the robot's hand to each grasp target for each task. 
The results are for a 1000 grasp trials performed for each object 
and each post-grasp intention performed using a model that does not
take as observation the object's category.}
\end{figure}

Figure \ref{fig:rewards_per_latdim}, shows the rewards of policy
with synergy models with increasing number of latent dimensions 
from 1 to 5, while Table \ref{table:success_rate_per_lat} shows
the corresponding success rates. Using one latent dimension achieves
a lower grasp success rate, probably due to the reduced synergy space 
that is unable to represent successful grasps. Larger than two
dimensions achieve similar grasp success rates, except the model
with 3 latent dimensions which is probably an experimental artifact.
Finally, in Table \ref{table:success_rate_wo_obj_cat}, we see
the success rate of the policy with and without the object category
as an observation. While the removal object category does not affect
the average grasp success rate, in Figure \ref{fig:distances_no_object_cat}
we see that the agent does not learn to grasp the object
from the correct location according to its post-grasp intention.
This is evident by the fact that for both  post-grasp intentions the
robot grasps the object closer to the handoff grasp target instead 
of grasping it from the corresponding target according to the task,
as shown in Figure \ref{fig:grasp_dists}.

\begin{table}[h]
\caption{Average success rate for each policy for 5000 trials.}
\label{table:success_rate}
\begin{center}
\begin{tabular}{ | m{3cm} | m{3cm} | }
\hline
\centering\textbf{Method} & 
\centering\arraybackslash\textbf{Success Rate}\\
\hline
\centering\arraybackslash Joint action space &  
\centering\arraybackslash 66\% \\
\hline
\centering\arraybackslash PCA action space & 
\centering\arraybackslash 71\% \\
\hline
\centering\arraybackslash VAE action space (ours) & 
\centering\arraybackslash \textbf{83\%} \\
\hline
\end{tabular}
\end{center}
\end{table}

\begin{table}[h]
\caption{Average success rate for policies with synergy models of different latent dimensions.}
\label{table:success_rate_per_lat}
\begin{center}
\begin{tabular}{ | m{3cm} | m{3cm} | }
\hline
\centering\textbf{Number of latent dimensions} & 
\centering\arraybackslash\textbf{Success Rate}\\
\hline
\centering\arraybackslash 1 &  
\centering\arraybackslash 59\% \\
\hline
\centering\arraybackslash 2 &  
\centering\arraybackslash 83\% \\
\hline
\centering\arraybackslash 3 &  
\centering\arraybackslash 68\% \\
\hline
\centering\arraybackslash 4 & 
\centering\arraybackslash 84\% \\
\hline
\centering\arraybackslash 5 & 
\centering\arraybackslash 81\% \\
\hline
\end{tabular}
\end{center}
\end{table}

\begin{table}[h]
\caption{Average success rate for each policy for 5000 trials.}
\label{table:success_rate_wo_obj_cat}
\begin{center}
\begin{tabular}{ | m{3cm} | m{3cm} | }
\hline
\centering\textbf{Method} & 
\centering\arraybackslash\textbf{Success Rate}\\
\hline
\centering\arraybackslash Policy without object category observation &  
\centering\arraybackslash 82\% \\
\hline
\centering\arraybackslash Policy with object category observation & 
\centering\arraybackslash 83\% \\
\hline
\end{tabular}
\end{center}
\end{table}

\begin{figure}[h] \centering
\includegraphics[width=0.35\textwidth]{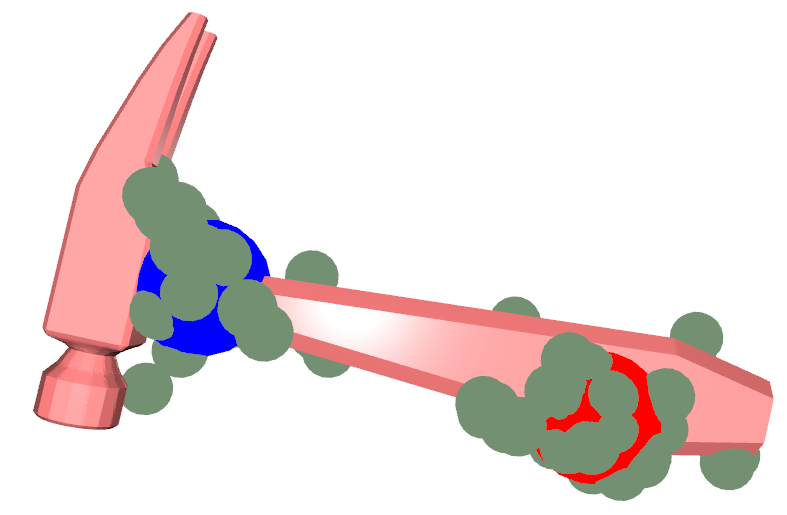}
\caption{\label{fig:hammer_clustering} 
Clustering results for grasp points on hammer object.}
\end{figure}

\begin{figure}[h] \centering
\includegraphics[width=0.49\textwidth]{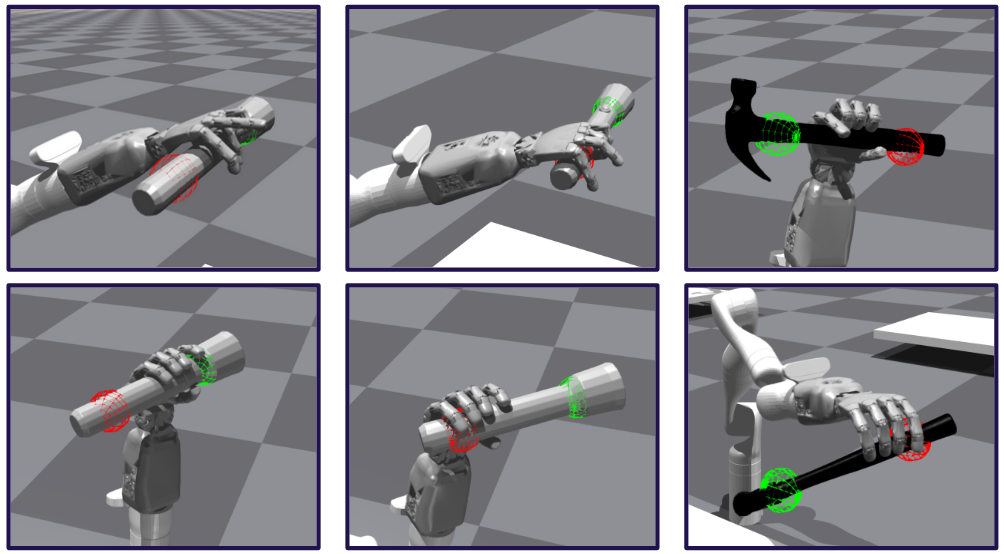}
\caption{\label{fig:grasps_examples} 
The first row of depicts grasps using the joint angles as
action space, while the second row depicts grasps from the policy 
that uses the synergy space as action space
}
\end{figure}

\textbf{Qualitative.}
A qualitative advantage of using the synergy space learned by the
VAE model as action space for the policy, is that the resulting 
grasps, i.e. the final finger configurations of the hand, resemble
the ones executed by the human subjects. The first row of Figure 
\ref{fig:grasps_examples} depicts grasps using the joint angles as
action space, while the second row depicts grasps from the policy 
that uses the synergy space as action space. In the first case, 
the grasp postures are unnatural, not all fingers are used and 
might become unstable for subsequent tasks. On the other hand,
the grasps from the synergy space are typical power grasps, 
similar to the ones the human subjects performed.

\section{CONCLUSION}
In summary, we have presented an agent that is able to grasp 
several objects while taking into account a post grasp intention.
The agent is trained on data collected from humans grasping objects
in order to use them and hand them off. The agent is trained 
using online reinforcement learning in a simulated environment.
We use the framework of postural synergies to reduce the action 
space in the reinforcement learning environment and generate human
like grasps. Our results demonstrate that the agent learns to 
grasp objects in a similar way that humans prefer to grasp them
according to the dataset. In addition, using a synergistic approach
we are able to improve the grasping success rate of the agent
compared to using directly the joint action space. Our work,
though still presents some limitations, such as that the agent
uses only the position of the grasping targets and not the orientation of
the object so in future work it would be useful to try to use the functional 
characteristics of the object. Finally, we assume that the
objects of a class, i.e. hammers, have similar sizes, and the policy
might fail on objects with very different size. Relying on vision data
for the observations, i.e. point clouds, might alleviate this limitation.


\bibliographystyle{IEEEtran}
\bibliography{references}

\vspace{12pt}
\end{document}